\pdfoutput=1

\documentclass[11pt]{article}

\usepackage[final]{acl}

\usepackage{times}
\usepackage{latexsym}

\usepackage[T1]{fontenc}

\usepackage[utf8]{inputenc}

\usepackage{microtype}

\usepackage{inconsolata}

\usepackage{graphicx}
\usepackage{multirow}
\usepackage{amsmath}

\title{Comparing representations of long clinical texts \\
for the task of patient note-identification}

\author{
\textbf{Safa Alsaidi}\textsuperscript{1}, 
\textbf{Marc Vincent}\textsuperscript{2}, 
\textbf{Olivia Boyer}\textsuperscript{3}, 
\textbf{Nicolas Garcelon}\textsuperscript{2}, \\ 
\textbf{Miguel Couceiro}\textsuperscript{4,5} and
\textbf{Adrien Coulet}\textsuperscript{1} \\ 
\\ 
\textsuperscript{1}Inria, Inserm, UPC, HeKA U1346, Paris, France \\ \texttt{\{safa.alsaidi, adrien.coulet\}@inria.fr}
\\ 
\textsuperscript{2}Data Science Platform, Imagine Institute, INSERM U1163, UPC, Paris, France \\ \texttt{\{marc.vincent, nicolas.garcelon\}@institutimagine.org} 
\\ 
\textsuperscript{3}Néphrologie Pédiatrique, Centre de Référence MARHEA, Hôpital Universitaire \\ Necker-Enfants Malades, 
Assistance Publique - Hôpitaux de Paris (APHP), \\
Institut Imagine, INSERM U1163, UPC, Paris, France 
\\ \texttt{olivia.boyer@aphp.fr} \\
\textsuperscript{4}Université de Lorraine, CNRS, LORIA, Nancy, France 
\\ 
\textsuperscript{5}INESC-ID, Instituto Superior Técnico, Universidade de Lisboa, Portugal 
\\ \texttt{miguel.couceiro@inesc-id.pt}\\ 
\textbf{Correspondence:} \texttt{safa.alsaidi@inria.fr} 
}

\begin{document}
\maketitle
\begin{abstract}

In this paper, we address the challenge of patient-note identification, which involves accurately matching an anonymized clinical note to its corresponding patient, represented by a set of related notes. This task has broad applications, including duplicate records detection and patient similarity analysis, which require robust patient-level representations. We explore various embedding methods, including Hierarchical Attention Networks (HAN), three-level Hierarchical Transformer Networks (HTN), LongFormer, and advanced BERT-based models, focusing on their ability to process medium-to-long clinical texts effectively. Additionally, we evaluate different pooling strategies (mean, max, and mean\_max) for aggregating word-level embeddings into patient-level representations and we examine the impact of sliding windows on model performance. Our results indicate that BERT-based embeddings outperform traditional and hierarchical models, particularly in processing lengthy clinical notes and capturing nuanced patient representations. Among the pooling strategies, mean\_max pooling consistently yields the best results, highlighting its ability to capture critical features from clinical notes. Furthermore, the reproduction of our results on both MIMIC dataset and Necker hospital data warehouse illustrates the generalizability of these approaches to real-world applications, emphasizing the importance of both embedding methods and aggregation strategies in optimizing patient-note identification and enhancing patient-level modeling.

\end{abstract}

\section{Introduction}

Representation learning focuses on learning compact, meaningful representations from raw data to make it easier for models to perform tasks such as classification, prediction, and clustering. In general, representation learning consists in learning dense representations, where complex, high-dimensional data are mapped to lower-dimensional spaces \citep{Liu2020RepresentationLA}. These representations capture underlying structure and essential features, preserving relevant information from the data. In the context of Natural Language Processing (NLP), representation learning has been widely applied and demonstrated impressive performance across various tasks, including downstream applications such as text classification, sentiment analysis, and machine translation \citep{glove:2014:pennington,Liu2020RepresentationLA,DBLP:journals/corr/abs-1904-03323}. 

Studies in healthcare have focused on learning patient representations from electronic health records (EHRs) to develop predictive models for patient outcomes, such as hospital readmissions, disease progression, or patient mortality rates \citep{Deo2015MachineLI,Zhu2015AligningBA,DBLP:conf/globecom/AuslanderGFBA20,Mahbub2022UnstructuredCN}. In recent years, EHRs have been widely adopted by many medical institutions, capturing comprehensive patient data throughout the care process \citep{Landi2020DeepRL,Escudi2018DeepRF,Steinberg2020LanguageMA,DBLP:conf/icml/LeM14}. Nonetheless, learning effective patient representations poses several challenges, one of which is determining what defines a ``good'' patient representation. The optimal representation can vary depending on the specific application, as well as factors such as data noise, missing values, and the type of data incorporated. For instance, representations designed for structured data \cite{Rajkomar2018ScalableAA} may differ significantly from those that incorporate both structured data and unstructured text \cite{DBLP:conf/acl/DeznabiIF21}. 
These challenges highlight the importance of investigating different representation learning methods to generate representations that are adapted not only to a specific task but also to the nature of the dataset.

In this paper, we address the task of patient-note identification, which consists in determining to which patient a particular note belongs. We focus exclusively on clinical texts, representing each patient as a set of chronologically ordered notes. While higher risks of patient-note mismatches have been reported with paper records, there is limited literature on this issue within modern EHR systems \citep{DBLP:journals/jamia/WilcoxCH11}, which further motivates our work. To this end, we investigate which text-based patient representation is best suited for the task of patient-note identification. 

Accurately identifying patient information is crucial in the medical field to ensure that a patient's medical history is up-to-date. Furthermore, this task has applications in biomedical informatics, including data cleaning and privacy-related tasks (e.g., assessing re-identification risk of patient data \citep{Lee2017ReidentificationOM}). 
More broadly, we believe that patient-note identification can serve as a foundational task for more advanced similarity-based tasks, such as clustering, diagnosing conditions by matching complex symptoms and medical histories, or finding "patients like mine" \citep{DBLP:journals/npjdm/GombarCCHS19, DBLP:journals/jbi/GarcelonNBSKSBH17}.

In this study, we conduct experiments on two datasets: MIMIC-III \citep{Goldberger2000PhysioBankPA} and an anonymized EHR dataset from our local hospital, the Necker hospital data warehouse (Dr. Warehouse) \citep{GARCELON201852}. We focus on the MIMIC-III dataset to develop and refine our approach to identify the best representation for the patient-note identification task, and only evaluate reproducibility of our findings using our local hospital dataset. We consider different embedding models to learn representations of potentially large sets of clinical notes associated with each patient, and evaluate and compare these representations by performing classification with traditional algorithms.

Our contributions are 3-fold:

\begin{itemize}
\item we clearly define the patient-note identification task and highlight its importance for studying patient representations;
\item we conduct an empirical comparison of patient representation methods for this task;
\item 
we attest that BERT-based model, using a sliding window mechanism and a combination of mean and max pooling, achieves the highest accuracy. 
\end{itemize}

\section{Related Work}

\subsection{Patient-Information Identification}

Despite the growing interest in patient-information identification, relatively few studies have explored this task using text, and to our knowledge, none have specifically addressed patient-note identification. This research gap further motivates our work.

Most efforts in patient matching or record linkage have focused on structured data \citep{Riplinger2020PatientIT}. Some prior studies have leveraged unstructured clinical text for patient-matching tasks. For example, \citet{Wornow2024ZeroShotCT} tackled the challenge of matching patients to clinical trials using a zero-shot LLM-based system. By evaluating unstructured clinical text against free-text trial criteria, their approach achieved state-of-the-art performance on the n2c2 2018 cohort selection benchmark. Clinician reviews indicated that the system provided coherent explanations for 97\% of correct decisions and 75\% of incorrect ones.

In contrast, other studies have explored deep learning approaches for patient identification using imaging data, particularly chest X-rays \citep{Ueda2023PatientIB, Packhuser2021DeepLP}. For instance, \citet{Packhuser2021DeepLP} trained a Siamese neural network to determine whether two frontal chest X-ray images belonged to the same patient, achieving an AUC of 0.9940 and a classification accuracy of 95.55\% on the ChestX-ray14 dataset. Similarly, \citet{Ueda2023PatientIB} proposed a deep metric learning approach using a deep convolutional neural network (DCNN) feature extractor and a classifier based on the cosine similarity index to verify patient identities from chest X-ray images. Their method achieved AUC values of 0.9999 and 0.9943 on the Morishita Laboratory and CheXpert datasets.

While these studies highlight the potential of deep learning for patient identification, our work fills a critical gap by focusing on text-based patient-note identification, an area that remains largely unexplored.

\subsection{Representation Learning}

Typically, EHRs comprise both structured (\textit{e.g.}, age, demographics, ICD codes, laboratory results) and unstructured data (\textit{e.g.}, free-text clinical notes such as radiology reports, discharge summaries, and medical images). The inherent complexity of EHRs has inspired numerous studies aimed at developing patient representations by learning optimized, dense numerical vectors 
\citep{Li2019BEHRTTF,DBLP:journals/jbi/SushilSLD18,Hashir2019TowardsUM,si2020patient}. 

Previous research has explored various approaches, 
including paragraph vectors \citep{DBLP:conf/icml/LeM14}, topic models \citep{Blei2001LatentDA}, word2vec embeddings \citep{DBLP:conf/nips/MikolovSCCD13}, and Hierarchical Attention Networks (HAN) \citep{si2020patient,Si2021ThreelevelHT}. For instance, \citet{DBLP:conf/globecom/AuslanderGFBA20} used word2vec and bag-of-words as feature extraction methods to learn patient representation from clinical notes for mortality prediction. 
\citet{DBLP:journals/jbi/SushilSLD18} learned generalized patient representations 
using a stacked denoising autoencoder and a paragraph vector model to predict patient mortality, primary diagnostic, procedural category, and patient gender. \citet{si2020patient} learned patient representations 
notes using a hierarchical attention-based recurrent neural network (HAN-RNN) with greedy segmentation and evaluated the model for mortality prediction and as a transfer learning pre-training model to downstream evaluation such as phenotype prediction. 

Representation learning from clinical texts, particularly using Bidirectional Encoder Representations from Transformers (BERT) \citep{DBLP:conf/naacl/DevlinCLT19}, has shown significant improvements in text-processing tasks like clinical named entity recognition (NER) and document classification \citep{DBLP:journals/corr/abs-1904-03323, DBLP:conf/bionlp/PengYL19, DBLP:journals/bioinformatics/LeeYKKKSK20}. 
BERT-based models have also been used for prediction in medicine. For example, \citet{Mahbub2022UnstructuredCN} used PubMedBERT to generate dynamic embeddings from clinical notes, enabling predictions of short-, mid-, and long-term mortality in adult ICU patients. However, due to BERT's 512-token limitation, longer clinical notes in these experiments had to be either truncated or split, which may have resulted in the loss of valuable context necessary for accurate predictions.

To address BERT's 512-token input limitation, models like BigBird \citep{Zaheer2020BigBT} and LongFormer \citep{Beltagy2020LongformerTL} have been employed to learn patient representations from longer clinical texts. These models support input sequences of up to 4,096 tokens (8 times the BERT limit), yielding substantial performance improvements in tasks such as long-text question answering and summarization. Additionally, \citep{Li2022ACS} introduced Clinical-Longformer and ClinicalBigBird, two pretrained language models specifically designed for lengthy clinical text processing. These models demonstrated superior performance in NER, question answering, and document classification tasks when handling lengthy documents.


These studies highlight the challenges involved in identifying the most suitable representation learning method for a specific task. In the case of patient-note identification, there is no one-size-fits-all solution, and the effectiveness of existing methods remains unclear, motivating empirical evaluation. 
In this work, we present an empirical comparison of four methods (HAN BERTLSTM, HTN, Longformer, and BERT) to assess their effectiveness in addressing the patient-note identification task. For BERT model, we introduce four different settings that explore different embedding strategies, using token embeddings (TE) or the [CLS] token, as well as applying a sliding window mechanism or restricting inputs to 512 tokens. This results in a total of seven experimental configurations.



\section{Datasets}

In this work, we use two distinct datasets of EHRs containing clinical notes: (1) the publicly available MIMIC-III dataset, which consists of ICU patient records in English, and (2) the Necker hospital data warehouse, containing French-language notes from nephrology patients. Below, we provide an overview of each dataset along with the preprocessing steps and selection criteria used to define our final cohorts.

\subsection{MIMIC-III}

MIMIC-III \citep{Johnson2016MIMICIIIAF} is a publicly available medical database that includes anonymized health records from 46,520 ICU patients treated at Beth Israel Deaconess Medical Center in Boston, Massachusetts, between 2001 and 2012. 
The MIMIC-III dataset provides a wide range of patient data, including demographics, vital signs, laboratory test results, clinical notes, and ICD-9 diagnosis codes. It contains 2,083,180 clinical notes across multiple categories, such as physician notes, nursing notes, discharge summaries, and radiology reports. The distribution of notes across different categories is shown in Table~\ref{tab:category_notes} in Appendix~\ref{sec:appendixa}. 

Firstly, we begin by performing several data cleaning operations: we exclude notes flagged as erroneous in MIMIC-III, those without a hospital admission identifier, notes lacking chart time information (\textit{i.e.}, the date and time the note was documented), and duplicated notes. For partially duplicated notes with identical chart times, we retain the longer note, ensuring that its text encompasses the content of the shorter note.

Secondly, we select only notes categorized as `Physician.' As presented in Table~\ref{tab:category_notes}, this category ranks fourth in terms of notes count and contains the longest notes, with an average length of 1,874 tokens and a median of 1,823 tokens. We hypothesize that these notes would provide the most comprehensive information about a patient's medical condition, enhancing the ability to accurately associate a clinical note with its corresponding patient. 

Thirdly, we remove outliers, using interquartile range (IQR) filtering and a threshold multiplier of 1.5, which results in excluding patients with more than 40 notes. As our work focuses on matching notes to the patient they belong to, we include only patients with at least two clinical notes: one serving as the target note for identification ($X_2$), and the other used to learn the patient’s representation ($X_1^i$). Ultimately, MIMIC-III dataset consists of 33,007 notes associated with 6,174 patients. The cohort design is illustrated in Figure~\ref{fig:cohort} in Appendix~\ref{sec:appendixa}. This dataset serves as a foundation for optimizing note representations for the task of patient-note identification.


\subsection{Necker Hospital Data Warehouse}

To assess the generalizability of our approach across different languages and medical specialties, we extended our analysis to EHRs from our local Necker hospital data warehouse (Dr. Warehouse), under IRB number \textit{2016–06-01}. This dataset encompasses a broad spectrum of clinical note types, such as consultations, hospitalization reports, discharge summaries, and laboratory results, spanning multiple departments. 

For our study, we focus on notes of nephrology patients hospitalized between 2018 and 2023. These selected notes have an average length of 1,237 tokens and a median of 897 tokens. We apply a similar preprocessing pipeline to the one used for MIMIC-III, filtering out note categories with limited text content, removing patients with exceptionally high note counts using IQR filtering and patients with less than two notes. All notes have been already pseudonymized. Ultimately, our dataset comprises 32,731 clinical notes associated with 5,145 patients.

Unlike MIMIC-III, which consists of English-language ICU patient notes, this dataset contains French-language notes from nephrology patients. This distinction allows us to evaluate the robustness of our approach across different languages and clinical settings.

\section{Methodology}
\subsection{Patient-note identification task} 
\label{sec:approach}

We define the patient-note identification task as a binary classification problem, where the input pair ($X_1^i$, $X_2$) maps to an output label $\hat{Y^i}$. Here, $X_1^i$ represents a unified representation of all notes belonging to patient $i$, excluding one randomly selected note ($X_2$), which is represented separately. $X_2$ denotes the representation of a single note, and $\hat{Y^i}$ is a binary label that equals 1 if $X_2$ belongs to patient $i$, and 0 otherwise. 

From an initial set $\mathcal{X} =\{ X^i_0\}$ with $i \in [1,n]$, $n$ the number of patients and $X^i_0$ is the set of notes associated with the patient $i$, we define our train and test sets as pairs $((X_1^i, X_2),Y^i) $. For each patient $i$, we designate randomly one clinical note $X_2$ as the \textit{target note}, while the remaining notes $X_1^i = X^i_0 \setminus X_2$ serve as the patient’s historical context (\textit{source notes}). 
%
To maintain a balanced representation between positive and negative examples, each randomly selected clinical note is associated once in our dataset to the correct patient, and once to a randomly chosen patient. Accordingly, the pair $(X_1^i, X_2)$ is either associated with $Y^i=1$ or $0$. This leads to a dataset with twice as many instances as patients. To guarantee consistency, patients are split in train and test sets before excluding $X2$, and building $(X_1^i, X_2)$ pairs. This ensures that both source and target notes of one patient are either in the train, or in the test set, avoiding data leakage. 


\subsection{Learning Patient-Note Representations} 
Successfully performing this task requires learning effective document-level representations. Consequently, our study evaluates the performance of various representation learning approaches. Drawing from previous work on document-level representation learning \cite{si2020patient, Liu2019LearningHR,Li2022ClinicalLongformerAC,Matondora2024NLPBP,Li2019BEHRTTF,Bazoge2024AdaptationOB}, we experiment with several models to evaluate their ability to generate effective representations for patient-note identification. Each model was selected to highlight distinct strategies for processing and aggregating clinical notes, including hierarchical approaches, transformer-based architectures, and hybrid designs that integrate both sequential and contextual information. Specifically, we experiment with a hierarchical attention network with BiLSTM and BERT at the word level (HAN BERTLSTM), a three-level hierarchical transformer network (HTN), Longformer, and BERT. Using these four models, we define seven different settings. For the BERT model, we consider two variants: one using token embeddings (TE) and the other using the [CLS] token (CLS). Token embeddings represent each individual token in the sequence, while the [CLS] token variant uses a special token at the beginning of the sequence to aggregate information for classification tasks. Additionally, we explore configurations both with and without a sliding window mechanism to address BERT’s 512-token limitation. The sliding window approach allows the model to process longer texts by splitting them into overlapping segments, whereas the alternative approach restricts inputs to a single 512-token sequence. Detailed descriptions of each model can be found in Appendix ~\ref{model}. The acronyms introduced in this section will be used consistently throughout the paper.

For clinical document representation, we evaluate and adapt several aggregation techniques traditionally used to transition from word-level to sentence-level and, subsequently, to document-level representations. These techniques include attention mechanisms, average pooling, max pooling, and mean\_max pooling \cite{DBLP:conf/acl/DeznabiIF21,Li2022ACS,Si2021ThreelevelHT,Mahbub2022UnstructuredCN}. To derive a single patient representation, we aggregate all note representations for a given patient into a unified representation using one of these four methods. 

\textit{Attention-based Aggregation (att)} employs a learnt attention mechanism to dynamically assign varying importance to each clinical note. \textit{Average Pooling or Mean Pooling (avg)} computes the mean representation of all clinical notes, capturing the overall feature distribution, while \textit{Max Pooling (max)} selects the highest value across note representations, emphasizing the most prominent features. Recent studies \cite{Si2021ThreelevelHT,Li2022ACS} suggest that \textit{Mean-max Pooling (mean\_max)}, which concatenates the average pooled and max pooled embeddings, often yields superior performance across predictive tasks by combining the strengths of both pooling strategies: the \textit{average} highlights overall feature distribution, while the \textit{max} emphasizes key dominant features.

To formalize this pooling strategy, let $\mathbf{r}_j$ be the vector representation of the $j$-th note of a given patient with $m$ notes. The aggregated patient representation $\mathbf{R}$ using mean\_max pooling is defined as:
\begin{equation*}
\resizebox{\columnwidth}{!}{ $\mathbf{R} = [\text{mean}(\mathbf{r}_1, \mathbf{r}_2, \ldots, \mathbf{r}_m) \oplus \text{max}(\mathbf{r}_1, \mathbf{r}_2, \ldots, \mathbf{r}_m)],$} 
\end{equation*}
where $\text{mean}(\cdot)$ computes the element-wise average, $\text{max}(\cdot)$ computes the element-wise maximum, and $\oplus$ denotes the concatenation operation.

Finally, these note representations serve as inputs to classifiers for the patient-note identification task. We evaluate five machine learning models: logistic regression (LR), random forest (RF), decision trees (DT), support vector machine (SVM), and XGBoost.
Performance of both the embedding methods and classifiers are measured with five key metrics: accuracy, precision, recall, F1-score, and area under the curve (AUC).

\section{Experiments and Results}


\begin{table*}[h!]
\centering
\resizebox{\linewidth}{!}{
\begin{tabular}{|c|c|c|c|c|c|}
\hline
\textbf{Dataset} & \textbf{Notes} & \textbf{Set} & \textbf{Token Count} & \textbf{Sentence Count} & \textbf{Vocabulary Size} \\
\hline
\multirow{4}{*}{MIMIC-III} & \textbf{Source Notes ($X_1^i$)} & Train & $8006.11 \pm 7416.22$ & $149.30 \pm 140.23$ & $15,510$ \\
& & Test & $8080.67 \pm 7503.55$ & $149.98 \pm 150.89$ & $14,050$ \\
\cline{2-6}

& \textbf{Target Note ($X_2$)} & Train & $1768.63 \pm 687.90$ & $33.88 \pm 26.31$ & $13,106$ \\
& & Test & $1782.04 \pm 706.93$ & $33.98 \pm 27.13$ & $11,609$ \\
\hline
\multirow{4}{*}{Necker Hospital Data Warehouse} & \textbf{Source Notes ($X_1^i$)} & Train & $8493.19 \pm 9935.22$ & $424.02 \pm 560.27$ & $17,335$ \\
 & & Test & $8578.87 \pm 9733.31$ & $427.30 \pm 545.52$ & $15,744$ \\
\cline{2-6} 
& \textbf{Target Note ($X_2$)} & Train & $1436.10 \pm 1013.69$ & $69.50 \pm 77.19$ & $14,907$ \\
& & Test & $1436.36 \pm 1033.57$ & $69.40 \pm 79.26$ & $12,962$ \\
\hline
\end{tabular}}

\caption{Mean number of tokens and sentences for the set of notes belonging to a single patient (source notes, $X_1^i$) and the target note ($X_2$) across train and test sets in both MIMIC-III and our local Necker hospital dataset, along with vocabulary sizes. Token count and vocabulary size are computed using the BERT WordPiece tokenizer. These values are computed over the three different train and test splits.}
\label{tab:combined}
\end{table*}

To ensure robustness, experiments were repeated three times on each dataset (MIMIC-III and Necker hospital dataset) with distinct random train and test splits, maintaining an 80/20 ratio. Table~\ref{tab:combined} provides details on the train and test sets, including the length of source ($X_1^i$) and target ($X_2$) notes, as well as the size of the associated dictionaries.

Table \ref{tab:random} presents the results of our experiments on MIMIC-III, keeping only results obtained for the best-performing classifier, which name is provided in the third column. The AUC score reflects the model’s ability to effectively distinguish between two classes: whether a note representation belongs to a given patient.

We observe that BERT\_TE\_sliding consistently outperforms all other models. Furthermore, mean\_max pooling consistently yields the best performance across all models and nearly all metrics as the aggregation method for patient representations. XGBoost also emerges as the top-performing machine learning algorithm across all models.


We evaluate the impact of pooling strategies (average, max, and mean\_max) on the performance of different models using paired t-tests to assess statistical significance. Mean\_max pooling outperforms mean and max pooling in most comparisons, with significant differences observed in most of the cases (p < 0.05). For hierarchical models, significant differences are observed between mean pooling and mean\_max pooling for both HAN BERTLSTM and HTN (p < 0.05). Additionally, max pooling shows a significant difference compared to mean\_max pooling for HTN (p < 0.05), but not for HAN BERTLSTM. 
Turning to the LongFormer model, mean pooling shows a significant difference compared to mean\_max pooling (p < 0.05), whereas no significant difference is found between max pooling and mean\_max pooling. Among BERT-based models, both BERT\_[CLS] and BERT\_TE exhibit significant differences between mean pooling and mean\_max pooling (p < 0.05). However, while max pooling differs significantly from mean\_max pooling for BERT\_TE (p < 0.05), no such difference is observed for BERT\_[CLS]. In sliding window approaches, significant differences emerge between mean pooling and mean\_max pooling for both BERT\_[CLS]\_sliding and BERT\_TE\_sliding (p < 0.05). Meanwhile, max pooling differs significantly from mean\_max pooling for BERT\_[CLS]\_sliding (p < 0.05), but not for BERT\_TE\_sliding.

To evaluate the generalizability of our results, we extend our analysis to EHRs from the Necker hospital data warehouse. For this experiment, we use only our best-performing model, BERT\_TE\_sliding, and test the three different aggregation methods to obtain patient-level representations. Since the dataset contains French clinical notes, we replace BERT with CamemBERT to accommodate the language difference. Results are obtained by conducting three independent runs and are reported in Table~\ref{tab:necker}. 
The results on the Necker hospital dataset show similar results to those of MIMIC-III, with the mean\_max aggregation method outperforming other pooling strategies. Statistical analysis using paired t-tests reveals a significant difference between mean\_max pooling and average pooling (p < 0.05), while the difference between mean\_max pooling and max pooling is not statistically significant. 

 \begin{table*}
\centering
\resizebox{\linewidth}{!}{
\begin{tabular}{|c|c|c|c|c|c|c|c|}
\hline
 \textbf{Model} & \textbf{Aggreg.} & \textbf{Classifier } & \textbf{Accuracy} & \textbf{Precision} & \textbf{Recall} & \textbf{F1} & \textbf{AUC}\\

 & & & (mean $\pm$ std.) & (mean $\pm$ std.) & (mean $\pm$ std.) & (mean $\pm$ std.) & (mean $\pm$ std.)\\

\hline

\multirow{4}{*}{\textbf{HAN BERTLSTM}} & att & RF & 0.64 $\pm$ 0.00 & 0.62 $\pm$ 0.00 & 0.69 $\pm$ 0.01 & 0.66 $\pm$ 0.01 & 0.70 $\pm$ 0.01\\


 & avg & SVM & 0.76 $\pm$ 0.00 & 0.82 $\pm$ 0.01 & 0.67 $\pm$ 0.01 & 0.73 $\pm$ 0.00 & 0.79 $\pm$ 0.00 \\

& max & XGBOOST & 0.75 $\pm$ 0.00 & 0.76 $\pm$ 0.00 & 0.74 $\pm$ 0.01 & 0.75 $\pm$ 0.01 & 0.82 $\pm$ 0.01 \\

 & mean\_max & XGBOOST & 0.76 $\pm$ 0.00 & 0.76 $\pm$ 0.01 & 0.75 $\pm$ 0.01 & 0.75 $\pm$ 0.00 & 0.83 $\pm$ 0.01 \\
\hline

\multirow{3}{*}{\textbf{3-level HTN}} & avg & XGBOOST & 0.74 $\pm$ 0.01 & 0.72 $\pm$ 0.01 & 0.80 $\pm$ 0.01 & 0.75 $\pm$ 0.01 & 0.82 $\pm$ 0.01 \\
& max & XGBOOST & 0.71 $\pm$ 0.01 & 0.68 $\pm$ 0.00 & 0.79 $\pm$ 0.01 & 0.73 $\pm$ 0.01 & 0.79 $\pm$ 0.01 \\
 & mean\_max & XGBOOST & 0.76 $\pm$ 0.01 & 0.74 $\pm$ 0.01 & 0.82 $\pm$ 0.01 & 0.77 $\pm$ 0.01 & 0.84 $\pm$ 0.00 \\
\hline

\multirow{3}{*}{\textbf{BERT\_TE}} & avg & XGBOOST & 0.85 $\pm$ 0.00 & 0.84 $\pm$ 0.00 & 0.86 $\pm$ 0.01 & 0.85 $\pm$ 0.00 & 0.93 $\pm$ 0.00 \\

& max & XGBOOST & 0.85 $\pm$ 0.01 & 0.86 $\pm$ 0.01 & 0.83 $\pm$ 0.01 & 0.85 $\pm$ 0.01 & 0.92 $\pm$ 0.01\\
 & mean\_max & XGBOOST & 0.87 $\pm$ 0.00 & 0.88 $\pm$ 0.00 & 0.85 $\pm$ 0.01 & 0.87 $\pm$ 0.00 & 0.94 $\pm$ 0.00 \\
 \hline

\multirow{3}{*}{\textbf{BERT\_[CLS]}} & avg & XGBOOST & 0.82 $\pm$ 0.00 & 0.81 $\pm$ 0.00 & 0.82 $\pm$ 0.01 & 0.82 $\pm$ 0.00 & 0.90 $\pm$ 0.00\\

& max & XGBOOST & 0.84 $\pm$ 0.00 & 0.83 $\pm$ 0.00 & 0.85 $\pm$ 0.01 & 0.84 $\pm$ 0.00 & 0.92 $\pm$ 0.00\\

& mean\_max & XGBOOST & 0.85$\pm$ 0.00 & 0.84 $\pm$ 0.01 & 0.85 $\pm$ 0.01 & 0.84 $\pm$ 0.01 & 0.93 $\pm$ 0.01\\
\hline
\multirow{3}{*}{\textbf{Longformer}} & avg & XGBOOST & 0.74 $\pm$ 0.01 & 0.74 $\pm$ 0.01 & 0.76 $\pm$ 0.02 & 0.75 $\pm$ 0.01 & 0.83 $\pm$ 0.00\\


& max & XGBOOST & 0.75 $\pm$ 0.01 & 0.75 $\pm$ 0.02 & 0.76 $\pm$ 0.02 & 0.75 $\pm$ 0.01 & 0.83 $\pm$ 0.00\\


& mean\_max & XGBOOST & 0.78 $\pm$ 0.01 & 0.78 $\pm$ 0.02 & 0.79 $\pm$ 0.02 & 0.78 $\pm$ 0.00 & 0.85 $\pm$ 0.00\\
\hline

\multirow{3}{*}{\textbf{BERT\_TE\_sliding}} & avg & XGBOOST & 0.85 $\pm$ 0.00 & 0.84 $\pm$ 0.01 & 0.87 $\pm$ 0.01 & 0.86 $\pm$ 0.00 & 0.94 $\pm$ 0.00 \\

& max & XGBOOST & 0.88 $\pm$ 0.00 & 0.89 $\pm$ 0.00 & 0.86 $\pm$ 0.01 & 0.88 $\pm$ 0.00 & 0.95 $\pm$ 0.00 \\


& \textbf{mean\_max} & \textbf{XGBOOST} & \textbf{0.90 $\pm$ 0.00} & \textbf{0.91 $\pm$ 0.00} & \textbf{0.88 $\pm$ 0.00} & \textbf{0.89 $\pm$ 0.00} & \textbf{0.96 $\pm$ 0.00}\\ 
\hline

\multirow{3}{*}{\textbf{BERT\_[CLS]\_sliding}} & avg & XGBOOST & 0.86 $\pm$ 0.00 & 0.85 $\pm$ 0.00 & 0.88 $\pm$ 0.00 & 0.87 $\pm$ 0.00 & 0.94 $\pm$ 0.00 \\

& max & XGBOOST & 0.85 $\pm$ 0.01 & 0.85 $\pm$ 0.01 & 0.85 $\pm$ 0.01 & 0.85 $\pm$ 0.01 & 0.93 $\pm$ 0.01 \\


& mean\_max & XGBOOST & 0.88 $\pm$ 0.00 & 0.88 $\pm$ 0.00 & 0.88 $\pm$ 0.00 & 0.88 $\pm$ 0.00 & 0.95 $\pm$ 0.00\\
\hline

\end{tabular}}

\caption{Best results reported based on AUC metrics across 4 models (7 different settings) among 5 different classification algorithms (LR, RF, SVM, DT, and XGBOOST), using MIMIC-III dataset. We report mean $\pm$ std. over 3 runs.}
\label{tab:random}
\end{table*}

\section{Discussion}

\begin{table*}[h!]
\centering
\resizebox{\linewidth}{!}{
\begin{tabular}{|c|c|c|c|c|c|c|c|}
\hline
 \textbf{Model} & \textbf{Aggreg.} & \textbf{Classifier} & \textbf{Accuracy} & \textbf{Precision} & \textbf{Recall} & \textbf{F1} & \textbf{AUC}\\

 & & & (mean $\pm$ std.) & (mean $\pm$ std.) & (mean $\pm$ std.) & (mean $\pm$ std.) & (mean $\pm$ std.)\\
 
\hline

\multirow{3}{*}{\shortstack{\textbf{BERT\_TE\_sliding} \\ (FR: CamemBERT)}} & avg & XGBOOST & 0.78 $\pm$ 0.01 & 0.79 $\pm$ 0.01 & 0.78 $\pm$ 0.02 & 0.78 $\pm$ 0.01 & 0.86 $\pm$ 0.01 \\

 & max & XGBOOST & 0.82 $\pm$ 0.00 & 0.83 $\pm$ 0.01 & 0.82 $\pm$ 0.02 & 0.83 $\pm$ 0.01 & 0.90 $\pm$ 0.01 \\


 & \textbf{mean\_max }& \textbf{XGBOOST} & \textbf{0.83 $\pm$ 0.01} & \textbf{0.84 $\pm$ 0.02} & \textbf{0.83 $\pm$ 0.00} & \textbf{0.83 $\pm$ 0.01} & \textbf{0.91 $\pm$ 0.01} \\

\hline

\end{tabular}}

\caption{Best results reported based on AUC metrics among 5 different classification algorithms (LR, RF, SVM, DT, and XGBOOST), using our local Necker hospital data warehouse. We report mean $\pm$ std. over 3 runs.}
\label{tab:necker}
\end{table*}

As mentioned previously, each experiment was conducted 3 times using a random sampling of train and test set. Although the reported standard deviation is small, this can be explained by the nature of our datatsets, \textit{i.e.}, our cohort selection. Given that our datasets consist of notes from specific categories (\textit{i.e.}, physician-only notes in the MIMIC-III dataset and nephrology-only notes in the Necker hospital dataset), which each tend to have similar language and structure within their respective categories, the model’s predictions are highly consistent. We believe this homogeneity within each category likely contributes to the low standard deviation observed. Despite this, our overall results (accuracy, AUC, and F1 score) indicate that our models effectively differentiate between notes and accurately matches them to their corresponding patient. 

In this version of the datasets, we conducted a single random drawing for each patient from their available set of clinical notes. However, to further expand the dataset, multiple random draws could be performed per patient, which would yield different patient representations.

The results obtained from our experiments emphasize the significance of model architecture, embedding strategies, and aggregation methods in optimizing performance for patient-note identification as shown in Table~\ref{tab:random}. Below, we discuss key observations and insights drawn from the performance metrics.
\begin{enumerate}
 \item {\bf Effect of Model Architecture:} The hierarchical models (HAN and HTN) demonstrated moderate performance.
 Among these, HAN BERTLSTM with mean\_max pooling achieved an F1 score of 0.75 and an AUC of 0.83. Similarly, the 3-level HTN model with mean\_max pooling achieved slightly better performance, with an F1 score of 0.77 and an AUC of 0.84, demonstrating the utility of hierarchical modeling. However, the overall performance of hierarchical models was surpassed by purely transformer-based models, including Longformer, which better captured contextual representations. 
 
 
 To elaborate, HAN and HTN rely on a fixed structure to aggregate information, which could limit their ability to detect nuanced relationships between sentences and words, particularly in long clinical notes. On the other hand, transformer-based models, such as Longformer and BERT, dynamically adjust word representations based on surrounding context. Given that we are working with clinical notes, we know that the meaning of terms could vary based on what follows and what precedes. Thus it is crucial to correctly identify or recognize the intended meaning of a particular term in a clinical note. While hierarchical models capture some structural patterns, they may miss more granular contextual cues, which are essential for accurately matching clinical notes to the correct patient.

 
 \item {\bf BERT Token Embedding (TE) vs. [CLS] Representations:} BERT models using TE achieved higher performance compared to those using [CLS] token representations. 
 While [CLS] embeddings are designed to encapsulate the overall sentence representation, their reliance on a single token representation might limit their ability to capture nuanced information spread across longer notes. In contrast, the token embeddings (TE) in BERT allows us to focus on the contextual representations of each token in a sequence. As demonstrated in the results, the mean\_max pooling strategy with BERT\_TE consistently yielded the best results, highlighting the effectiveness of combining token embeddings with contextual attention mechanisms in capturing fine-grained details from clinical notes.

 \item {\bf Longformer vs. BERT Sliding Window:} The Longformer model addresses BERT's token-length limitation by processing up to 4096 tokens, outperforming hierarchical models but falling short of BERT's sliding window configurations. Longformer achieved an F1 score of 0.83 and an AUC of 0.92, demonstrating its capability to handle lengthy clinical notes. 
 In contrast, BERT\_TE\_sliding with mean\_max pooling achieved the highest performance, with an F1 score of 0.89 and an AUC of 0.96. This success highlights the sliding window approach's ability to capture contextual information distributed across long notes. By employing overlapping windows, the model attended to diverse parts of the notes while maintaining contextual integrity. This method proved to be superior to Longformer’s fixed sliding window attention mechanism, as it enabled chunk-specific embeddings to be aggregated effectively.
 \item {\bf Pooling strategies:} mean\_max pooling consistently yielded the best results, likely due to its ability to capture both global and localized features across embeddings. By focusing on the maximum values, max pooling reduces the influence of less relevant or noisy features and, at the same time, ensures that the most important features are prominently represented in the final patient-level representation. In contrast, average pooling calculates the mean across all clinical note representations to derive the final patient representation, which can result in the loss of critical information, particularly when vital details are scattered across notes.

\end{enumerate}

In addition to the findings on the MIMIC-III dataset, the results on the Necker hospital dataset highlight two key points. First, our model demonstrates strong adaptability to a different dataset, effectively addressing the task of patient-note identification. Second, the results on our local dataset align with our previous experiments on the MIMIC-III dataset, where the mean\_max aggregation method generally outperforms other pooling strategies or performs similarly in a few cases, where no significant difference was observed compared to max pooling. These results highlight the versatility of our approach, demonstrating its effectiveness across diverse datasets and languages.

\section{Conclusion} \label{sec:conclusion}

Patient-note identification is a fundamental problem in the domain of medical informatics. While not extensively explored, the risks associated with patient-note mismatches can have serious consequences, particularly in ICU settings. In this work, we developed a framework to address this challenge using unstructured clinical notes from the MIMIC-III database. We evaluated various embedding models (HAN BERTLSTM, HTN, Longformer, and BERT) and aggregation methods (average, max, and mean\_max pooling) to generate patient-level representations. Our findings highlight that transformer-based models with advanced aggregation strategies, such as mean\_max pooling combined with a sliding window approach, are highly effective for capturing fine-grained contextual information and ensuring accurate patient-note identification. Additionally, experiments on an external dataset validated the generalizability of our approach. By adapting to French clinical notes with CamemBERT, the model maintained strong performance, demonstrating its robustness across diverse datasets and settings.


\section{Limitations}

Through this work, we emphasize the importance of patient-note identification and the potential of leveraging raw clinical notes for predictive modeling. While our approach shows strong performance, it is not without limitations. Our first limit lies in the use of generic language models to learn patient representations rather than using domain-specific architectures. Future research could explore more specialized models, such as ClinicalMamba \cite{Yang2024ClinicalMambaAG} and ModernBERT \citep{warner2024smarterbetterfasterlonger}, and investigate alternative aggregation strategies. These approaches may enhance representation quality and help mitigate potential information loss inherent in processing complex clinical text. However, it is important to consider the potential biases embedded in the pretraining data of these models, as such biases can impact both the generalizability and fairness of their application in clinical settings.

Another limit lies in the exclusive focus on unstructured clinical notes within our current framework. Integrating structured data, such as laboratory results or vital signs, alongside unstructured text could yield more comprehensive patient representations and allow for more nuanced comparative analyses.

Additionally, while we validated our approach using an external dataset, we did not assess its effectiveness on downstream clinical tasks, such as predictive modeling or forecasting, where clinical notes serve as primary or supplementary input. Such evaluations could offer further insights into the practical utility of the learned patient representations. 

Finally, benchmarking our method against large language models (LLMs), including ChatGPT or GPT-4o, could provide valuable perspectives for assessing the scalability, accuracy, and overall effectiveness of our approach in the context of patient-note identification.

\section*{Acknowledgments}

Experiments presented in this paper were carried out using computational clusters equipped with GPU from CLEPS infrastructure from the Inria of Paris\footnote{\url{https://paris cluster-2019.gitlabpages.inria.fr/cleps/cleps userguide/architecture/architecture.html}}. 

This work was 
supported by the French ANR\footnote{Agence Nationale de la Recherche.} project 
 ``Analogies: from theory to tools and applications'' (AT2TA), ANR-22-CE23-0023.
Additional support was provided by ANR project C'IL-LICO, ANR-17-RHUS-0002, and the European Union (Horizon-health-2022-disease-06-two stage, grant 101080717). The C'IL-LICO project and study protocol received approval from the French National Ethics and Scientific Committee for Research, Studies and Evaluations in the Field of Health (CESREES) under the number \#2201437. The data processing was approved by the French Data Protection Authority (CNIL) with a waiver of informed consent under number $DR-2023-017//920398v1$.


\bibliography{custom}

\appendix

\section{Appendix}

\subsection{Notes Statistics}
\label{sec:appendixa}

\begin{table}[h!]
\centering
\begin{tabular}{|l|r|}
\hline
\textbf{CATEGORY} & \textbf{NUMBER OF NOTES} \\ \hline
Nursing/other & 822,497 \\ \hline
Radiology & 522,279 \\ \hline
Nursing & 223,556 \\ \hline
ECG & 209,051 \\ \hline
Physician & 141,624 \\ \hline
Discharge summary & 59,652 \\ \hline
Echo & 45,794 \\ \hline
Respiratory & 31,739 \\ \hline
Nutrition & 9,418 \\ \hline
General & 8,301 \\ \hline
Rehab Services & 5,431 \\ \hline
Social Work & 2,670 \\ \hline
Case Management & 967 \\ \hline
Pharmacy & 103 \\ \hline
Consult & 98 \\ \hline
\end{tabular}
\caption{Number of notes per category in the MIMIC-III dataset.}
\label{tab:category_notes}
\end{table}

\begin{figure}[h!]
 \includegraphics[width=\columnwidth]{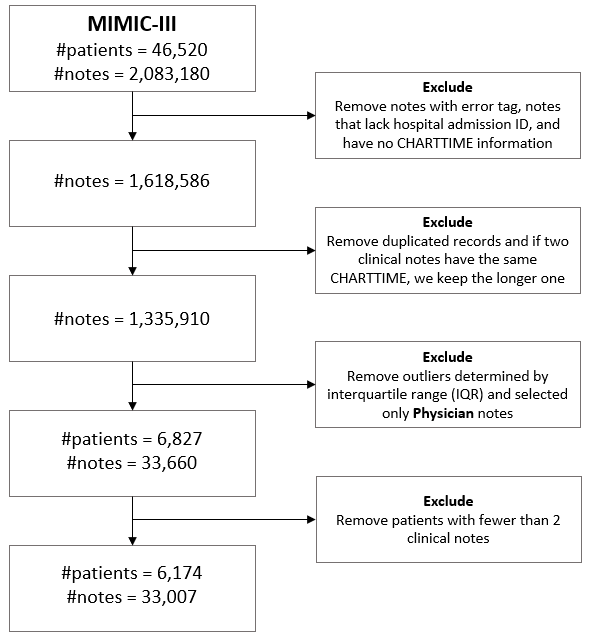}
 \caption{Cohort design, MIMIC-III dataset.}
 \label{fig:cohort}
\end{figure}




\subsection{Models Details} \label{model}

\subsubsection{HAN BERTLSTM} 
Following the architecture proposed by \citep{Si2021ThreelevelHT}, we adapted their HAN BiLSTM model\footnote{Model code is available at \url{https://github.com/Yuqi92/3-level-HTN-MIMIC.git}} to our task. The model integrates a BERT component as a fully trainable word-level encoder, followed by BiLSTMs and a pooling strategy to hierarchically learn sentence-level and document-level embeddings. The BiLSTMs and a global context-based attention mechanism capture sequential information at both the sentence and document levels, while a pooling strategy aggregates embeddings from one level to the next, extracting salient features at each stage.

In our implementation, we employed the BERTBase model at the word level. BERTBase comprises 12 layers, 768 hidden units, and 12 attention heads. It is pretrained on general-domain text datasets, including English Wikipedia (2.5 billion words) and the BookCorpus dataset \citep{Zhu2015AligningBA} (800 million words). The model uses the WordPiece tokenizer \citep{DBLP:journals/corr/WuSCLNMKCGMKSJL16} and has an input token limit of 512.

\begin{figure}[h!]
 \includegraphics[width=\columnwidth]{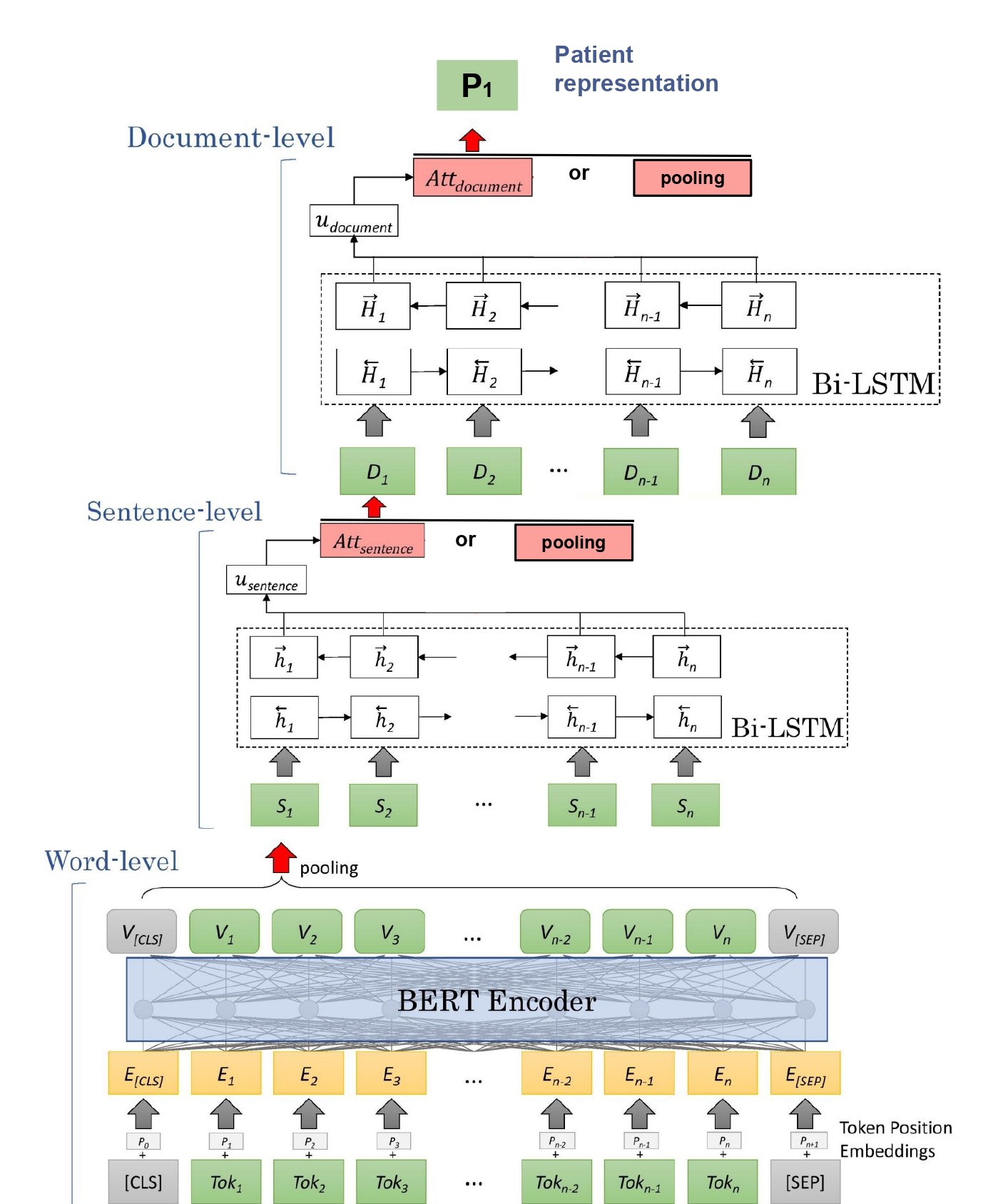}
 \caption{Overview of the HAN architecture incorporating BERT and BiLSTM with attention or pooling strategies for hierarchical aggregation. Adapted from \citep{Si2021ThreelevelHT}.}
 \label{fig:HAN}
\end{figure}

To generate word-level embeddings, we applied either the attention mechanism resulting from the original HAN BERTLSTM pre-training or one of several pooling strategies, namely average pooling, max pooling, or mean\_max pooling, to the BERT output. These word-level embeddings were then passed through the BiLSTM encoder to capture sentence-level features, where the same attention or pooling strategies were applied to produce final sentence embeddings. Similarly, document-level embeddings for individual clinical notes were obtained by applying the same strategies at the next hierarchical level. For patient-level representation, we aggregated the embeddings of all notes associated with a single patient. This was achieved using either an attention mechanism or a pooling strategy. Experimenting with these various pooling strategies allowed us to assess their impact on the patient-note identification task. The architecture of the model is shown in Figure~\ref{fig:HAN}. 

\subsubsection{HTN} 
\begin{figure}[h!]
 \includegraphics[width=\columnwidth]{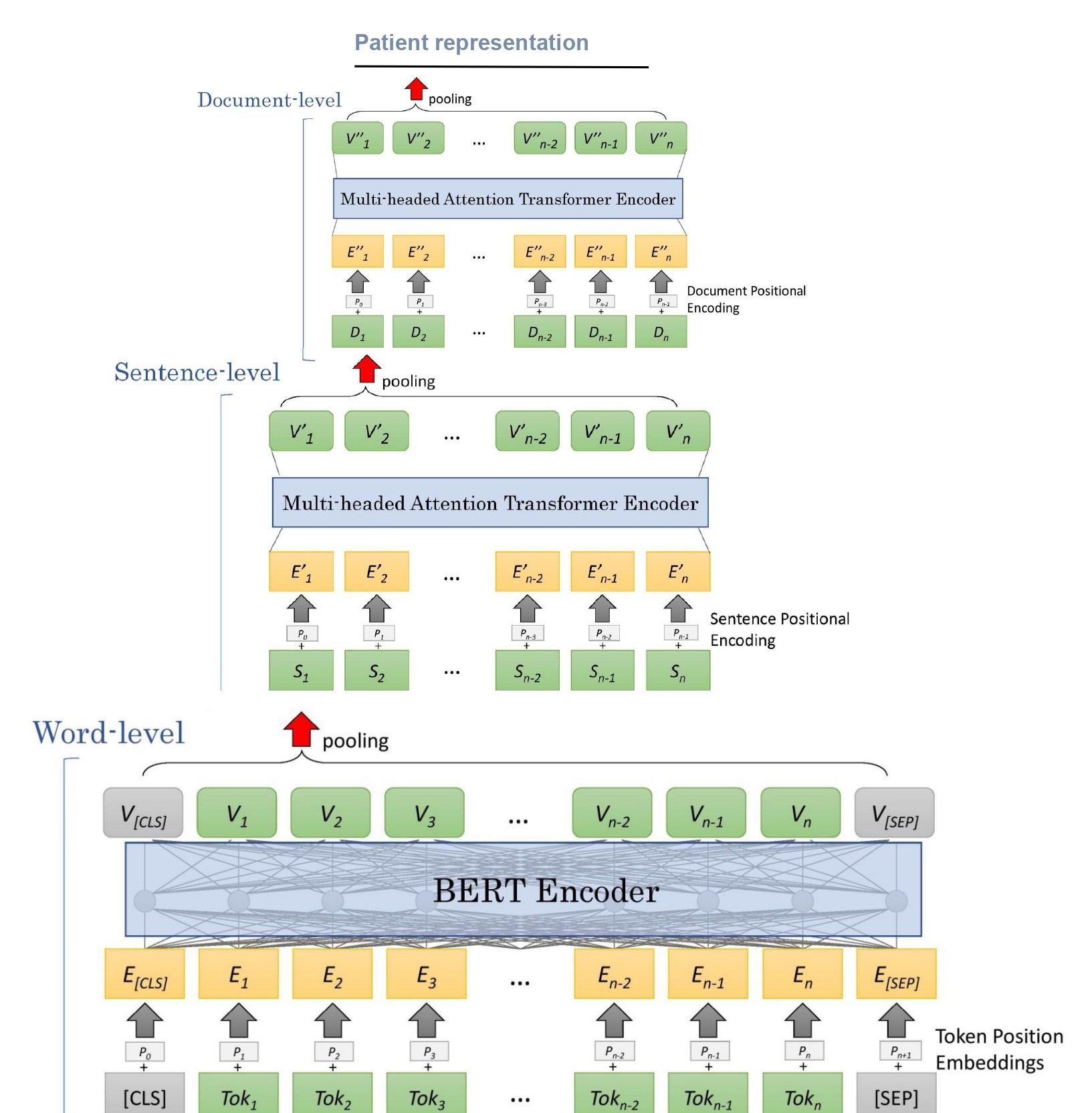}
 \caption{Overview of the HTN architecture incorporating BERT and Multi-head Transformer Encoder with pooling strategies for hierarchical aggregation. Adapted from \citep{Si2021ThreelevelHT}.}
 \label{fig:HTN}
\end{figure}

As our second model, we evaluated the three-level Hierarchical Transformer Network (HTN) \footnote{Model code is available at \url{https://github.com/Yuqi92/3-level-HTN-MIMIC.git}}, proposed by \citep{Si2021ThreelevelHT} and illustrated in Figure~\ref{fig:HTN}. The model architecture progressively constructs representations from the word level to the document level. At the word level, the model integrates a BERT encoder, experimenting with different BERT variants to balance model size and sequence length. At the sentence and document levels, it employs a Transformer-based encoder architecture inspired by \citep{Vaswani2017AttentionIA}, using multiheaded self-attention to identify key features and pooling to condense representations for the next level. Inputs are cropped or padded to fixed sizes at all levels (word, sentence, document). Further details about the model can be found in \citep{Si2021ThreelevelHT}.

For our experiments, we used the BERTBase model at the word level. To construct higher-level representations from word to document level, we experimented with three pooling strategies: average, max, and mean\_max pooling. Patient-level representations were then derived by aggregating note-level representations for each patient using the same pooling strategies.

\subsubsection{Bert-based Models} As our third model, we aimed to evaluate the standalone performance of BERT \cite{DBLP:conf/naacl/DevlinCLT19}, a widely used transformer model, to establish a robust baseline. This experiment was designed to understand the capability of BERT in capturing semantic and contextual information from clinical notes without leveraging additional hierarchical mechanisms or pretrained domain-specific adaptations. BERT has proven to be highly effective in various NLP tasks, making it a strong candidate for text representation in this context. Although specialized models like ClinicalBERT \cite{Huang2019ClinicalBERTMC} have shown strong results in clinical applications, we opted not to use them to avoid potential bias. ClinicalBERT is pretrained on the MIMIC-III dataset, which overlaps with our experimental data, potentially confounding the evaluation. By employing the generic BERTBase model, we ensure a fairer evaluation of our approach.

\paragraph{Token Embeddings Representations}

In the first approach, the final representations are derived from token embeddings in the text. Clinical notes are first split into sentences, and each sentence is tokenized. The tokenized input is passed into BERTBase, which generates embeddings for each token in the sentences. To obtain a single vector representation of a sentence, we pool the token embeddings using one of three strategies: average, max, or mean\_max pooling, represented as:
\begin{equation}
 \label{eq:eq2}
\resizebox{\columnwidth}{!}{
 $S_{repr} = avg/max/mean\_max(TE(w_1), TE(w_2), \ldots, TE(w_n))$}
\end{equation}
, where $S_{repr}$ refers to the sentence representation and TE($w_n$) is the token embedding representation of each token in the sentence. 
 To construct document-level embeddings, the sentence embeddings are appended and aggregated using the same pooling strategies (average, max, or mean\_max) along the dimension of sentence: 
 \begin{equation}
 \label{eq:eq3}
\resizebox{\columnwidth}{!}{
 ${N}_{repr} = avg/max/mean\_max(S_1 repr, S_2 repr, \ldots, S_n repr)$}
\end{equation}
Finally, for patient-level representation, where each patient has a set of clinical notes, the document-level embeddings are aggregated using average, max, or mean\_max pooling. This results in a single vector representation that captures the information from all notes associated with the patient.

\paragraph{[CLS] Token-Based Representation}

In the second approach, instead of learning token embeddings and aggregating them to obtain sentence representations, we directly extract the [CLS] token representation for each sentence. For each sentence, the sum of all token embeddings is passed through the Transformer layers (TL) to compute the final representation of the [CLS] token: 
 \begin{equation}
 \label{eq:eq4}
\resizebox{\columnwidth}{!}{
$[[CLS]]_{repr} = TL(TE(w_1)+ TE(w_2)+ \ldots + TE(w_n)$}
\end{equation} 
The [CLS] representations for all sentences are concatenated to form the input for document-level embedding. To obtain the document-level embeddings, we apply the same aggregation strategies (average, max, or mean\_max) across the [CLS] token representations of sentences:\\
 \begin{equation*}
\resizebox{\columnwidth}{!}{
${N}_{repr} = avg/max/mean\_max([[CLS]]_1 repr, [[CLS]]_2 repr, \ldots, [[CLS]]_n repr)$}
\end{equation*} 
 
For patient-level representation, document-level embeddings from all notes associated with a patient are further aggregated using the same pooling strategies (average, max, or mean\_max), producing a single vector representation for the patient.

By experimenting with these two methods, we aim to comprehensively evaluate BERT's effectiveness at capturing representations at sentence, document, and patient levels, while establishing a strong comparative baseline for this task.

\subsection{LongFormer}

\begin{figure}[h!]
 \includegraphics[width=\columnwidth]{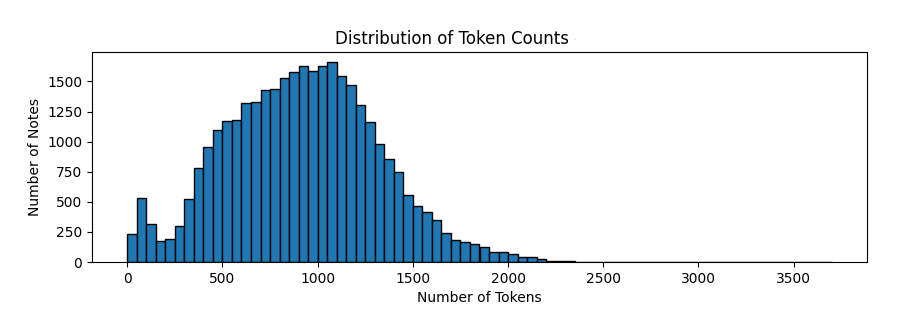}
 \caption{Number of tokens across MIMIC-III Physician notes.}
 \label{fig:token}
\end{figure}

Upon reviewing the MIMIC-III Physician notes, we observed that 28,266 out of 33,660 notes exceed 512 tokens, indicating that approximately 84\% of the notes exceed the token limit imposed by BERT. This suggests that the 512-token constraint may restrict the amount of information BERT can effectively capture. Figure~\ref{fig:token} illustrates the token distribution across the Physician clinical notes in the MIMIC-III dataset. Given this limitation, we sought to explore a model that could handle longer sequences more effectively, motivating our decision to experiment with Longformer as our fourth model. Unlike BERT, Longformer can process sequences up to 4096 tokens, addressing BERT's token constraint. It does this through a sliding window attention mechanism by allowing each token to attend only to a fixed window of neighboring tokens. Additionally, Longformer incorporates a global attention mechanism for selected tokens, such as the [CLS] token, enabling the model to capture broader context in longer documents.


Longformer is pretrained on a mix of general-purpose datasets, including scientific and news articles, designed to handle long-form text. 

\subsection{Bert-based [CLS] Token or Token Embedding (TE) With Sliding Window Model}

Motivated by the Longformer model, we conducted a new set of experiments with BERT, implementing the sliding window mechanism for both token embedding (TE) and [CLS] token-based representations. We believe this approach (our fifth model) will not only overcome the token limitation imposed by BERT but also enable the model to focus on different parts of a clinical note, often spread across various sections, thus minimizing information loss typically associated with pooling.

\begin{figure}[h!]
 \includegraphics[width=\columnwidth]{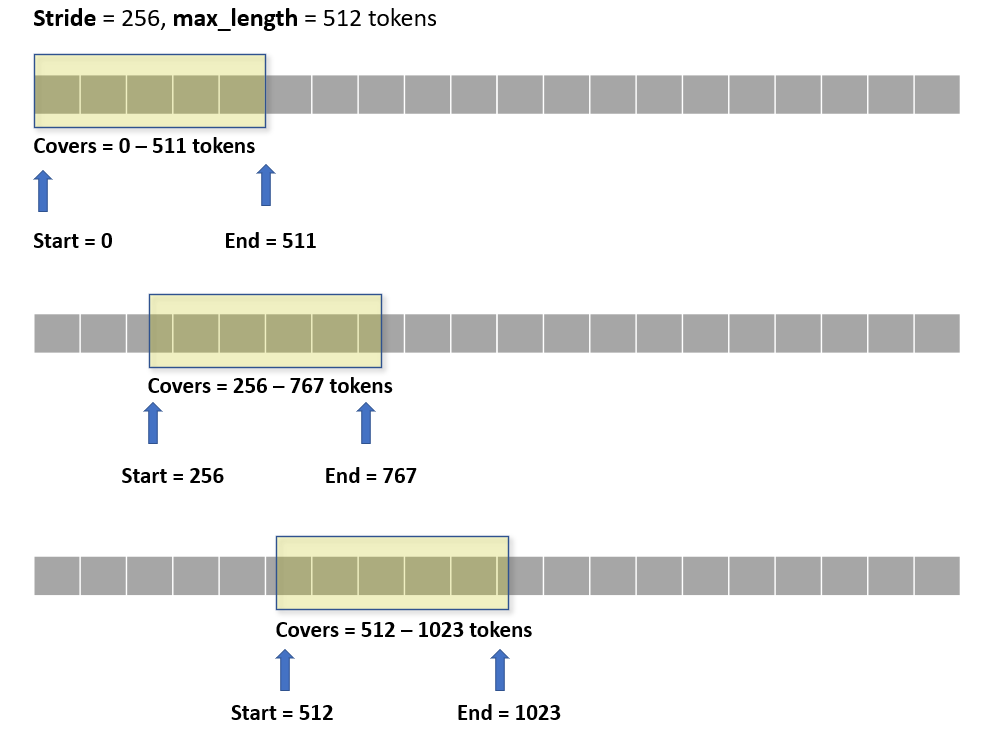}
 \caption{Sliding window approach.}
 \label{fig:sliding}
\end{figure}

To implement this, we begin by splitting each note into individual sentences. If a sentence exceeds 512 tokens, a sliding window is applied. The window starts at an initial position and processes a chunk of the sentence up to 512 tokens. It then moves by a specified stride and processes the next chunk. For our experiments, we set the stride value to 256 tokens, meaning that each window overlaps with the next one. We believe this overlapping strategy helps preserve contextual information when learning embeddings, as shown in Figure~\ref{fig:sliding}.

This process continues until the entire sentence is covered. For each chunk, we obtain sentence-level embeddings using either the token embedding (TE) or [CLS] representation, depending on the approach being tested. To illustrate, consider a note consisting of two sentences: one long sentence containing more than 512 tokens and a shorter sentence with exactly 512 tokens. The long sentence ($S_1$) can be represented as $S_1 = S_{repr 1.1}, S_{repr 1.2}, S_{repr 1.3}$, and the short sentence ($S_2$) as $S_2 = S_{repr 2}$. The final document-level embedding for the entire note is then computed as the average of all sentence embeddings:\\
 \begin{equation*}
\resizebox{\columnwidth}{!}{
${N}_{repr} = avg/max/mean\_max(S_{repr 1.1}, S_{repr 1.2}, S_{repr 1.3}, S_{repr 2})$}
\end{equation*}

It is important to note that our sliding window approach differs from the one used in Longformer. While Longformer processes the entire input using a global attention mechanism and a sliding window to select specific tokens to attend to, our approach divides the text into chunks, ensuring that each part of the sentence is processed separately before combining them into a final representation. This chunk-based approach allows us to handle very long sentences in a more structured manner.

\end{document}